\documentclass[conference,a4paper]{APSIPA2023}
\usepackage{multirow}
\usepackage{graphicx}
\usepackage{amsmath}
\usepackage[psamsfonts]{amssymb}
\usepackage{amsxtra}
\usepackage{threeparttable}
\usepackage{subfigure}

\usepackage{geometry}
\usepackage{fancyhdr}

\fancypagestyle{firststyle} {
    \fancyhf{}
    \fancyhead[C]{2023 Asia Pacific Signal and Information Processing Association Annual Summit and Conference (APSIPA ASC)}

}

\begin{document}

\title{Transformer-based Image Compression with Variable Image Quality Objectives}

\author{%
\authorblockN{%
Chia-Hao Kao$^{\star}$, Yi-Hsin Chen$^{\star}$, Cheng Chien, Wei-Chen Chiu, and Wen-Hsiao Peng
}
\authorblockA{%
National Yang Ming Chiao Tung University, Taiwan}
}

\maketitle
\def\thefootnote{$\star$}\footnotetext{Equal contribution.}
\thispagestyle{firststyle}

\begin{abstract}
This paper presents a Transformer-based image compression system that allows for a variable image quality objective according to the user's preference. Optimizing a learned codec for different quality objectives leads to reconstructed images with varying visual characteristics. Our method provides the user with the flexibility to choose a trade-off between two image quality objectives using a single, shared model. Motivated by the success of prompt-tuning techniques, we introduce prompt tokens to condition our Transformer-based autoencoder. These prompt tokens are generated adaptively based on the user's preference and input image through learning a prompt generation network. Extensive experiments on commonly used quality metrics demonstrate the effectiveness of our method in adapting the encoding and/or decoding processes to a variable quality objective. While offering the additional flexibility, our proposed method performs comparably to the single-objective methods in terms of rate-distortion performance.

\end{abstract}

\section{Introduction}
Compared to traditional image codecs, learned image codecs can be easily adapted to any differentiable quality metrics. When a learned image codec is trained separately with different quality metrics, the decoded images may exhibit distinctive visual characteristics. For instance, involving mean squared error (MSE) in the training objective usually leads to blurry reconstructed images, particularly at low bitrates. On the other hand, when trained to minimize the Learned Perceptual Image Patch Similarity (LPIPS)~\cite{zhang2018unreasonable}, the codec typically yields visually pleasing images, but may sometimes introduce undesirable artificial patterns~\cite{mentzer2020high}. 
By supporting multiple quality objectives, a compression system can cater to a wide range of application requirements and user preferences. It enables the user to adapt the quality objective to their needs, whether it is to maximize the perceptual quality or reconstruction fidelity of the decoded image, or a weighted combination of these objectives. Therefore, developing a flexible compression system that can support multiple quality objectives is of importance in meeting the diverse demands of different image compression applications.

Although end-to-end learned image compression has made significant progress~\cite{cheng2020learned, he2022elic, zhu2022transformer}, most existing methods have to train separate models for different quality objectives. Normally, each model is optimized for one specific quality metric, which can be MSE, multi-scale structural similarity (MS-SSIM), or a weighted combination of these metrics. However, this approach is impractical in terms of computational resources and storage requirements. 



Recently, Agustsson~\emph{et al.}~\cite{agustsson2022multi} propose a compression system with a conditional decoder for adaptive reconstruction of images to trade off high fidelity (i.e. low MSE) for realism. From the same bitstream, it is able to decode an image with variable quality objectives. 
However, they~\cite{agustsson2022multi} only address the trade-off between realism and MSE, disregarding other metrics. Furthermore, their conditioning mechanism is performed on the decoder side only, without exploring the potential benefits of incorporating it into both the encoder and decoder.


\begin{figure}[t]
    \centering
    \includegraphics[width=0.43\textwidth]{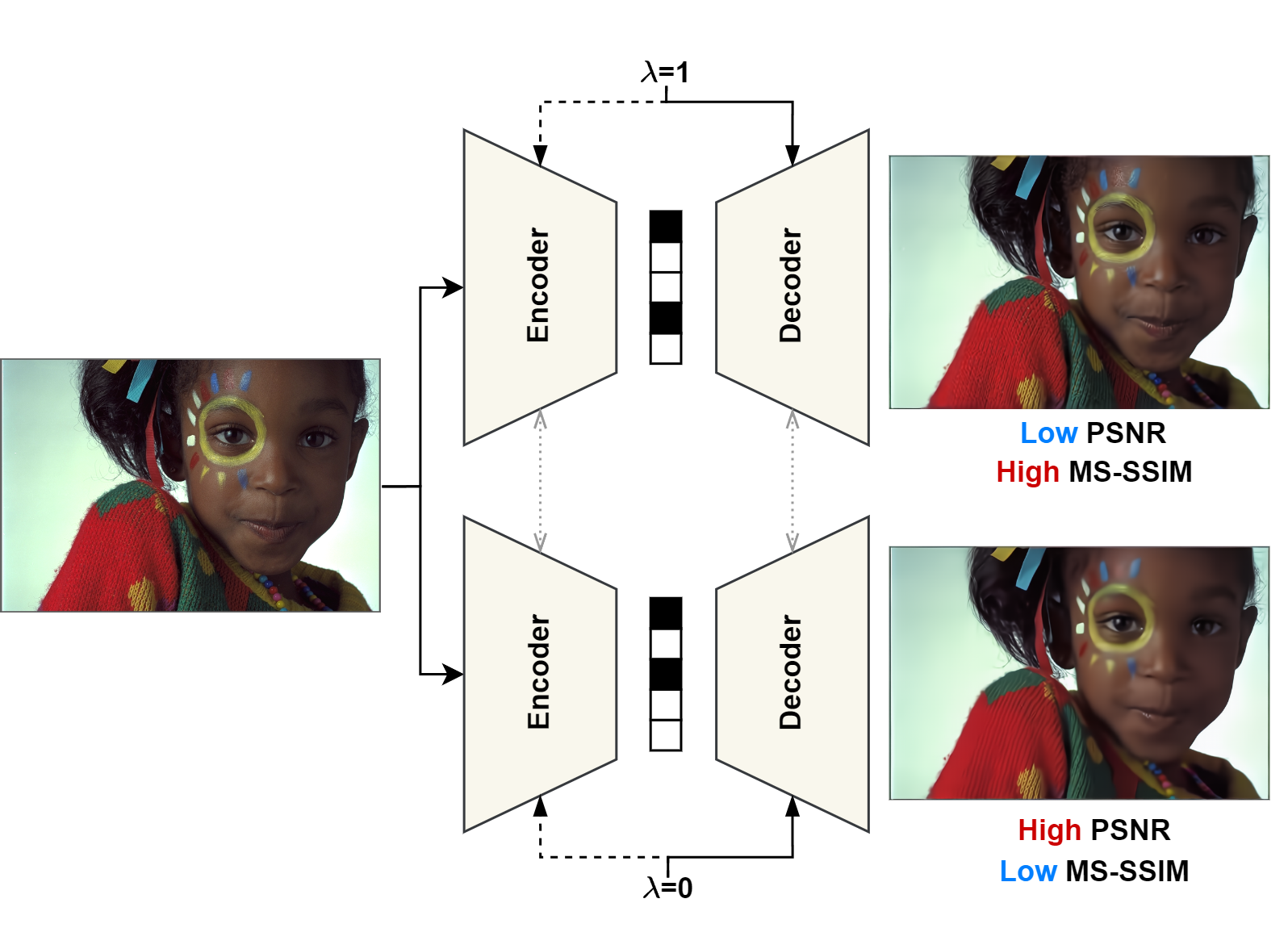}
    \caption{Illustration of our proposed framework. By adjusting the condition parameter $\lambda$, the reconstructed images exhibit distinct characteristics, enabling customizable quality balance based on specific preferences. } 
    
    \label{fig:teaser}
    \vspace{-3mm}
\end{figure}

In this work, we propose a Transformer-based learned image codec that is able to trade off one image quality objective for another (e.g. Peak-Signal-to-Noise-Ratio versus MS-SSIM). Transformer-based image compression systems~\cite{lu2022transformer, zhu2022transformer, liu2023learned} have attracted lots of attention recently for their comparable or even superior coding efficiency to CNN-based methods. Fig.~\ref{fig:teaser} illustrates the main idea of our framework, which is designed to adapt the encoding and/or decoding processes to optimize the decoded image quality according to a given trade-off between two selected quality objectives. Motivated by the recent success of prompt tuning techniques~\cite{liu2021prompt, jia2022visual}, we propose a prompt generation network to adapt the coding process according to the specified quality objective. Specifically, we explore two variations. The first applies prompt tuning to both encoder and decoder, specializing the bitstream for a specific quality objective. In comparison, the second adapts only the decoder, allowing a single bitstream to be decoded into images that meet variable quality objectives.

The contributions of this paper are as follows:
\begin{itemize}
    \item We introduce a Transformer-based image compression framework that is able to adapt the decoded image quality to a variable objective.
    \item To our best knowledge, this is the first work that leverages prompt-tuning techniques to trade off one image quality objective for another. Additionally, we explore two variants of the proposed method for different application scenarios.
    \item Extensive experiments confirm the effectiveness of our proposed method. Specifically, when both encoder and decoder are adapted, our method is able to trade off one quality objective for another effectively while achieving similar rate-distortion performance to the case when the codec is optimized solely for one of the two objectives. 
\end{itemize}






\section{Related Works}

\begin{figure*}
\centering
\includegraphics[width=0.95\textwidth]{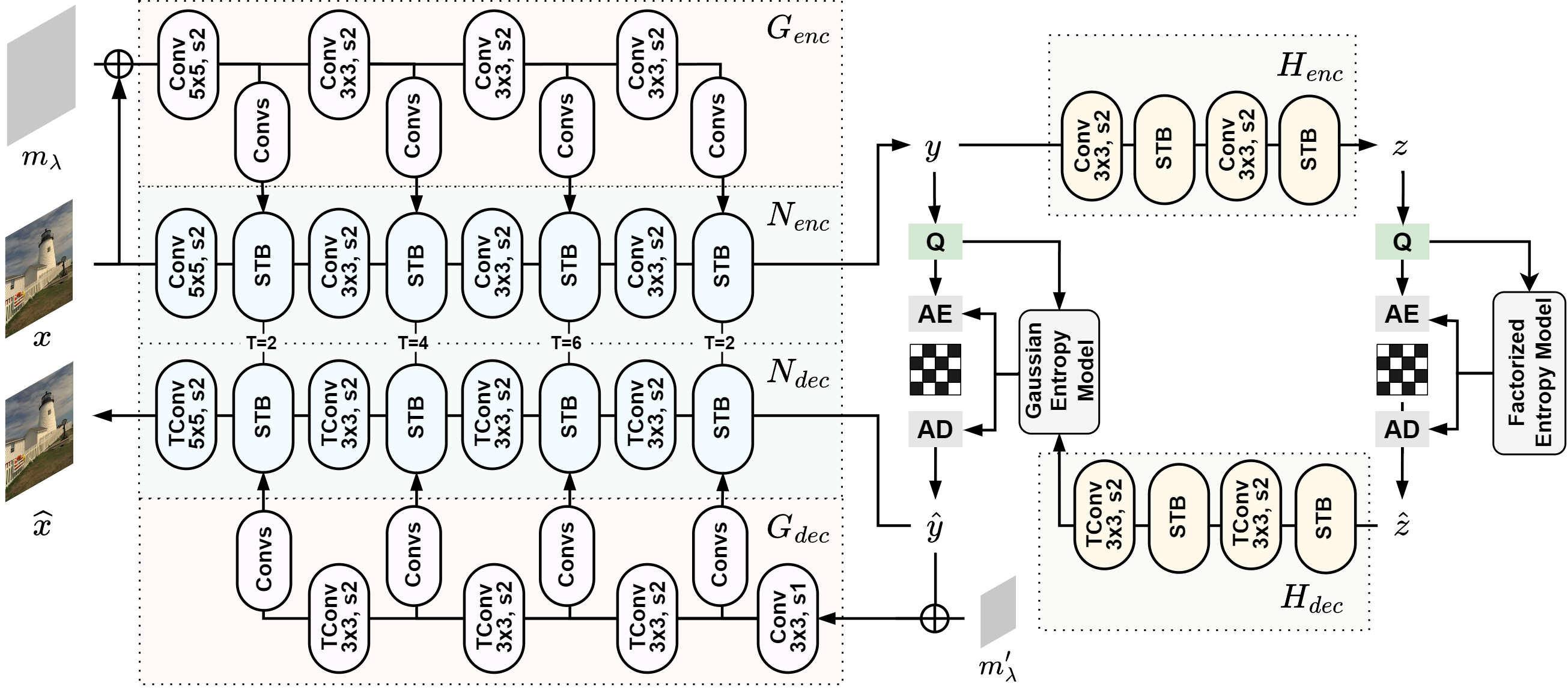}
\caption{Overall architecture of our proposed method. We introduce prompt generation networks $G_{enc}, G_{dec}$ to generate prompt tokens as additional input for adapting to variable image quality objectives.}
\label{fig:arch}
\vspace{-3mm}
\end{figure*}

\subsection{Learned Image Compression}
End-to-end learned image compression methods~\cite{balle2018variational, minnen2018joint, cheng2020learned, he2022elic} have made significant amount of progress over the past few years, even outperforming traditional codecs, such as intra coding in HEVC~\cite{HEVC} and VVC~\cite{VVC}. While most of the existing literature in this field adopt CNN-based autoencoder as their foundation, Transformer-based compression systems~\cite{zhu2022transformer, lu2022transformer} have emerged as promising alternatives, particularly for their low computational complexity and outstanding coding efficiency. Zhu~\emph{et al.}~\cite{zhu2022transformer} adopt Swin-Transformer~\cite{liu2021swin} architecture for image compression, showing comparable performance to its CNN-based counterparts with significantly reduced decoder-side kMACs (multiply-accumulate-operations). Similarly, Lu~\emph{et al.}~\cite{lu2022transformer} utilize Swin-Transformer blocks but substitute the patch merging/splitting operations with more general convolution and transposed convolution layers. More recently, Liu~\emph{et al.}~\cite{liu2023learned} propose a mixed Transformer-CNN image compression system that further improves the coding efficiency.

\subsection{Conditional Compression Systems}

Conditional compression systems designs are commonly adopted for variable rate coding, enabling a single compression model to encode input images at various bit-rates based on the given trade-off of rate and distortion. In this line of works, the condition is often applied through altering the distribution of the feature maps. Cui~\emph{et al.}~\cite{cui2021asymmetric} introduce gain units that perform channel-wise scaling on the image latents for continuous rate adaptation. Song~\emph{et al.}~\cite{song2021variable} use spatial feature transform (SFT) layers~\cite{wang2018recovering} to perform element-wise affine transformation on the intermediate feature maps of the autoencoder, achieving both variable rate and ROI coding simultaneously. In a similar vein, Wang~\emph{et al.}~\cite{wang2022block} target Transformer-based models and perform channel-wise scaling on value matrices in all self-attention layers for variable rate control.
Recently, Agustsson~\emph{et al.}~\cite{agustsson2022multi} extend the idea to trade off realism for fidelity (i.e. MSE) of reconstructed images, introducing a conditional decoder that performs channel-wise shifting on the intermediate feature maps inside each residual blocks. The condition is applied to adapt the decoder for different levels of realism and produce the reconstructed images accordingly. However, this method only addresses the trade-off between realism and fidelity, and limits the adoption of conditioning mechanism on the decoder side alone. 

\subsection{Prompt Tuning}
Prompting techniques~\cite{liu2021prompt, lester2021power, liliang2021prefix} were first introduced in natural language processing (NLP) field for efficiently adapting large pre-trained Transformer models to various downstream tasks. The main idea is to keep the entire large pre-trained Transformer frozen, and introduce additional input tokens, known as prompts, for adaptation. These tokens are involved in the self-attention and act as a guidance for the current task. Recently, Jia ~\emph{et al.}~\cite{jia2022visual} extend the idea to Vision Transformers~\cite{dosovitskiy2020vit} and computer vision tasks. They prepend learnable prompts to the image tokens at each Transformer blocks, achieving comparable or even superior performance than fine-tuning the entire model on downstream task while only learning less than 1\% of the total parameters.

\section{Proposed Method}

\subsection{Preliminary}
Most of the learning-based image compression models~\cite{balle2018variational, minnen2018joint, cheng2020learned, he2022elic} share a similar coding architecture, consisting of a main autoencoder and a hyperprior autoencoder~\cite{balle2018variational}. As illustrated in Fig.~\ref{fig:arch}, our framework also adopts this design. 
Specifically, given an input image $x\in\mathbb{R}^{3\times H\times W}$ of height $H$ and width $W$, a learned non-linear analysis transform $N_{enc}$ encodes the image into its latent representation $y\in\mathbb{R}^{192\times \frac{H}{16}\times \frac{W}{16}}$. To efficiently transmit $y$, it is first uniformly quantized into $\hat{y}$, which is then entropy coded using a learned prior distribution $p(\hat{y})$. This prior distribution is modeled using the hyperprior autoencoder in a content-adaptive manner~\cite{minnen2018joint}. As shown, the hyper-analysis transform $H_{enc}$ converts the quantized latent $\hat{y}$ into the side information $\hat{z}\in\mathbb{R}^{128\times \frac{H}{64}\times \frac{W}{64}}$ and the quantized $\hat{z}$ is transmitted for generating the learned prior distribution for entropy coding $\hat{y}$. Lastly, the synthesis transform $N_{dec}$ decodes $\hat{y}$ from the bitstream to obtain the reconstructed image $\hat{x}\in\mathbb{R}^{3\times H\times W}$.


\subsection{System Overview}
Our work proposes an image compression system capable of producing reconstructed images that reflect variable image quality objectives. Specifically, given two selected quality objectives, the coding process is adapted to trade off one objective for the other.
Fig.~\ref{fig:arch} illustrates the overall framework of our proposed method, which is built upon TIC~\cite{lu2022transformer}, a Transformer-based image compression system. For the analysis transform $N_{enc}$, synthesis transform $N_{dec}$, and hyper-autoencoder $H_{enc}, H_{dec}$ of TIC, the commonly-seen CNN-based design is replaced with interleaving convolutional layers and Swin-Transformer blocks (STB). Note that we discard the context model in favor of a simpler Gaussian prior for entropy coding. To achieve variable image quality objectives, our system incorporates an additional input, a uniform lambda map $m_\lambda\in\mathbb{R}^{1\times H\times W}$ of the same resolution as the input image on the encoder side, and a smaller-scale uniform lambda map $m'_\lambda\in\mathbb{R}^{1\times \frac{H}{16}\times \frac{W}{16}}$ on the decoder side. Both $m_\lambda$ and $m'_\lambda$ are populated with the same parameter $\lambda\in[0,1]$. By adjusting the value of $\lambda\in[0,1]$, users can specify their desired trade-off between the two quality objectives. At the two extreme ends, where $\lambda=0$ or $\lambda=1$, the system reduces to simply optimizing for one specific quality objective. Inspired by the success of prompt tuning~\cite{jia2022visual}, we leverage the idea of prompting to condition our Swin-Transformer based codec for adaptation. \textcolor{black}{In this work, we explore two variants of our conditioning method for two different application scenarios. The first variant incorporates the conditioning mechanism into both the encode and decoder, specializing the resulting bitstream for a specific quality objective. The other variant adopts only the decoder-side adaptation, enabling a single compressed bitstream to be decoded into images with variable quality objectives. The details of these variants are presented in the last paragraph of Section~\ref{sec:prompt}.}

\subsection{Prompt-based Conditioning}
\label{sec:prompt}
Drawing inspiration from prompt tuning methods~\cite{jia2022visual}, we employ additional learnable tokens, termed prompts, to condition our Transformer-based compression model for variable image quality objectives. We introduce prompt generation networks $G_{enc}$ and $G_{dec}$ as add-on modules on top of our main autoencoder. 
As shown in Fig.~\ref{fig:arch}, $G_{enc}$, which consists of several convolutional layers, is introduced to the encoder. It takes as input the concatenation of the input image $x$ and lambda map $m_\lambda$ to generate prompt tokens for each STB in $N_{enc}$. 
Similarly, the decoder incorporates $G_{dec}$, which shares the same architecture as $G_{enc}$ but replaces convolutions with transposed convolutions for upsampling. The input to $G_{dec}$ is the concatenation of the quantized latent $\hat{y}$ and a smaller-resolution lambda map $m_\lambda'$. What sets our method apart from~\cite{jia2022visual}, in which prompts are directly learned parameters, is the introduction of prompt generation networks for generating prompt tokens adaptively according to the given $\lambda$s and the input image or latent.

\begin{figure}[t]
    \centering
    \includegraphics[width=0.42\textwidth]{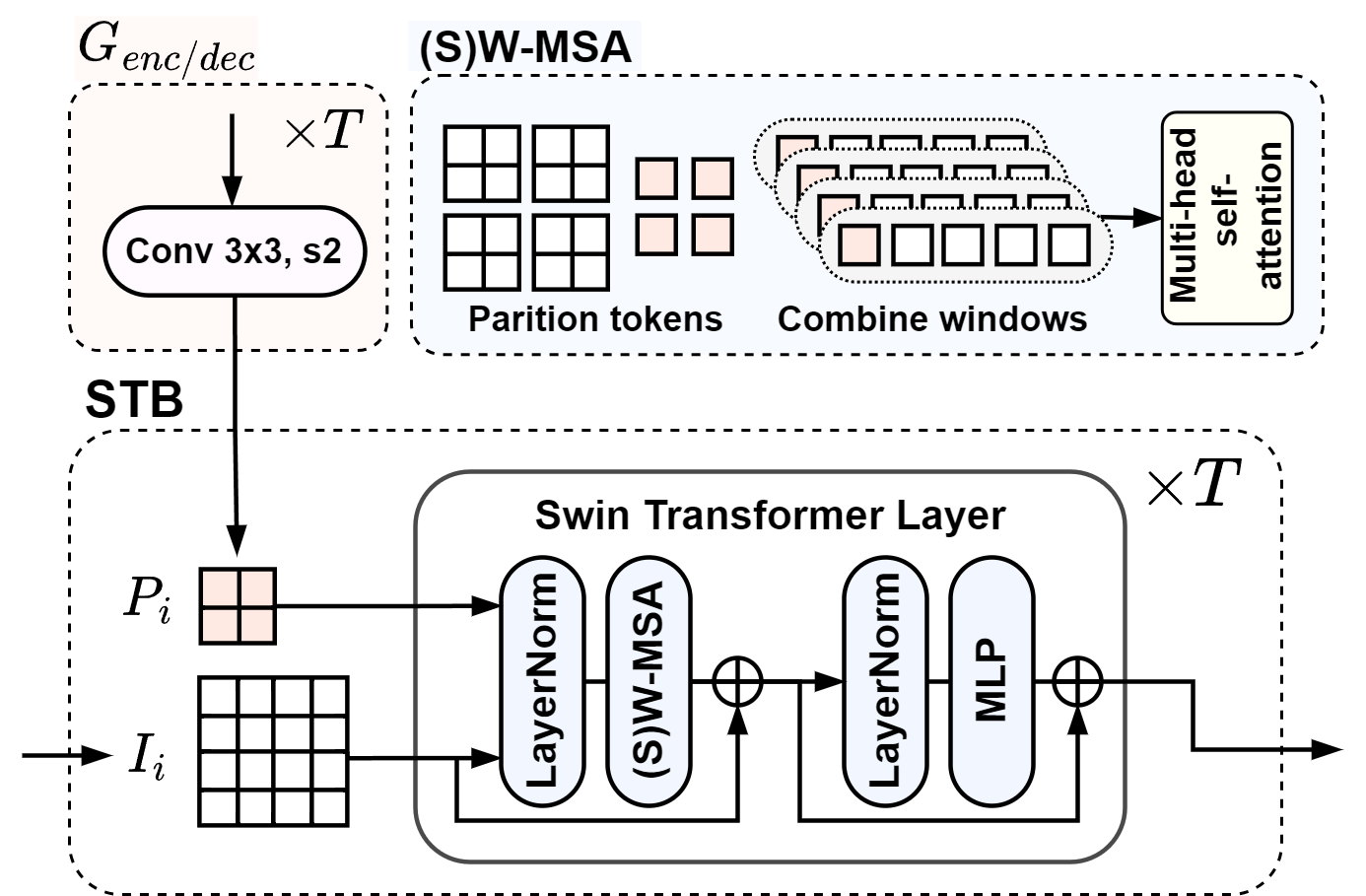}
    \caption{Detailed architecture of Swin-Transformer Block (STB).}
    \label{fig:stb}
    \vspace{-4mm}
\end{figure}


Fig.~\ref{fig:stb} illustrates the structure of STB and how we incorporate prompt tokens. A regular STB is comprised of multiple Swin-Transformer layers, with $T$ denoting \textcolor{black}{the number of layers} and the image tokens $I_i$ being the input to the $i^{th}$ layer. Each Swin-Transformer layer consists of layer normalization, window-based multi-head self-attention (W-MSA), and a MLP layer to aggregate neighboring information. Note that the window for W-MSA is shifted every other layer for tokens across windows to exchange information. To perform prompting, the input to $i^{th}$ Swin-Transformer layer additionally includes a set of prompt tokens $P_i$ obtained through the prompt generation network. In our design, the number of tokens for $P_i$ is only a quarter of that of $I_i$ for reducing computational complexity. 

\begin{figure}[t]
\centering
\subfigure[LPIPS]{
\centering
\includegraphics[width=0.475\linewidth]{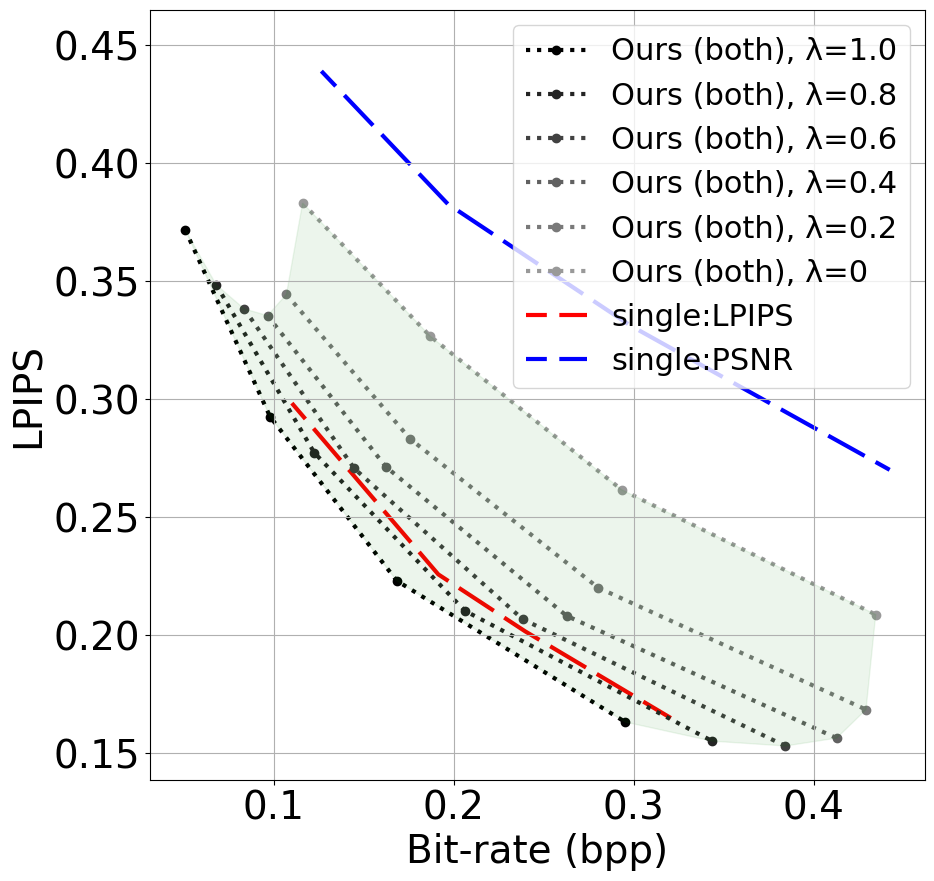}
}
\subfigure[PSNR]{
\centering
\includegraphics[width=0.455\linewidth]{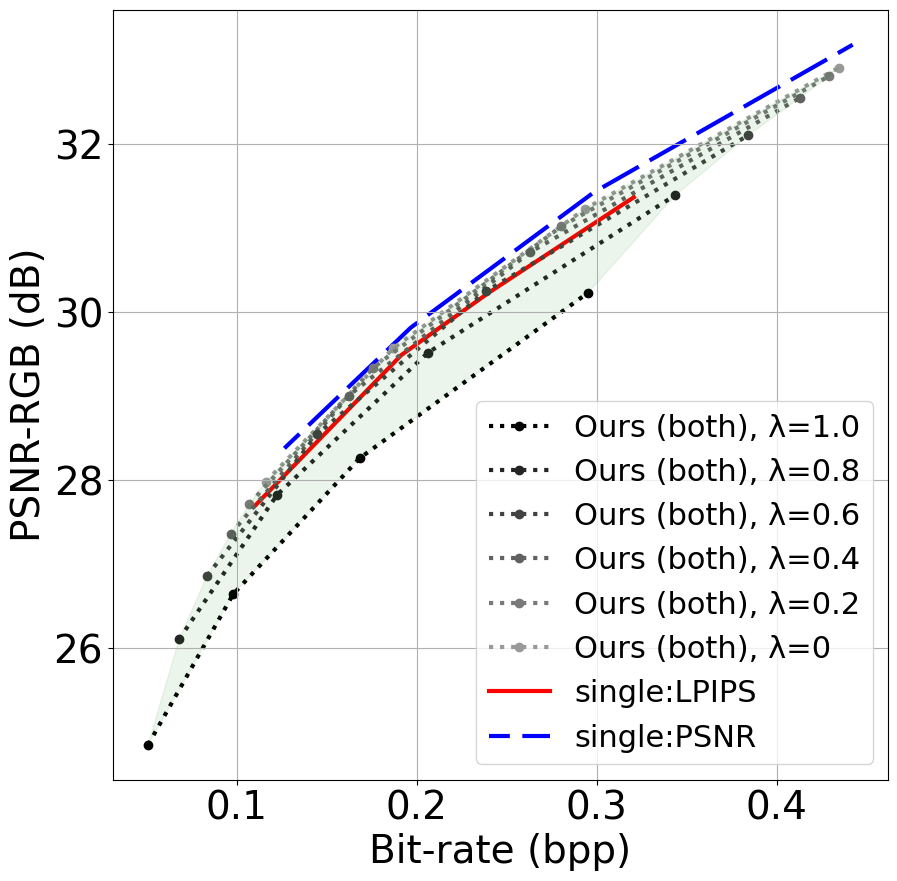}
}
\caption{Rate-distortion performance under (LPIPS, PSNR) pair with both-side prompts. Green shaded area indicates the adaptive range of our method.}
\label{fig:RD_main_both_LPIPS}
\end{figure}
\begin{figure}[t]
\centering
\subfigure[MS-SSIM]{
\centering
\includegraphics[width=0.46\linewidth]{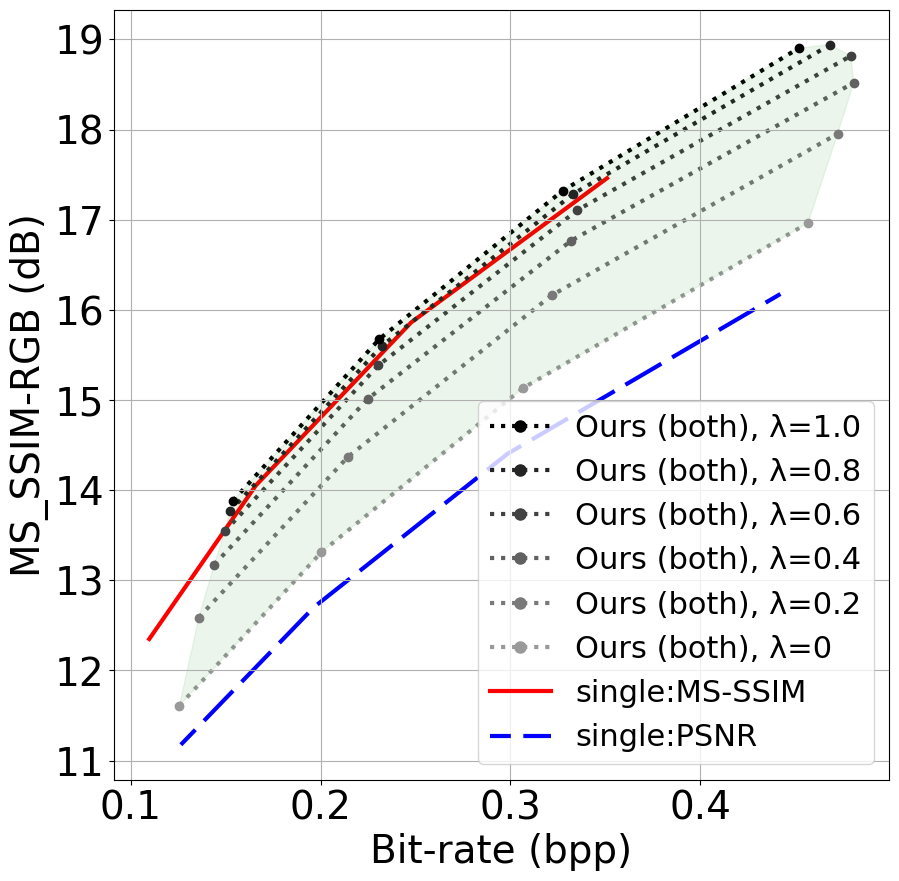}
}
\subfigure[PSNR]{
\centering
\includegraphics[width=0.46\linewidth]{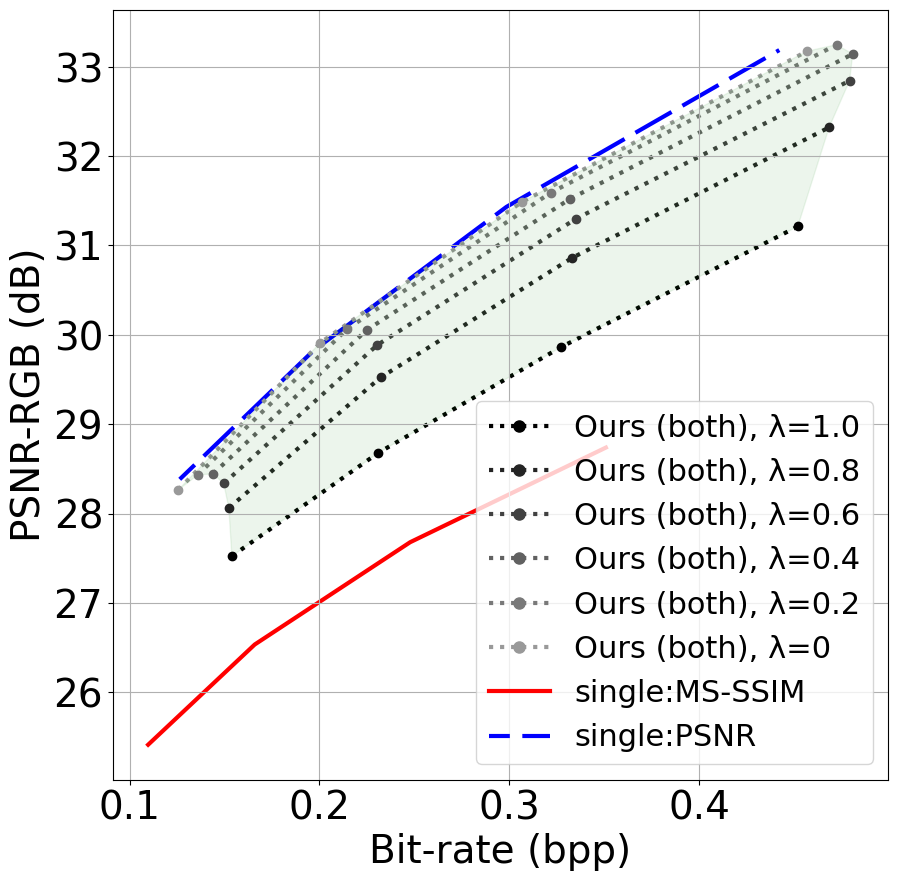}
}
\caption{Rate-distortion performance under (MS-SSIM, PSNR) pair with both-side prompts. Green shaded area indicates the adaptive range of our method.}
\label{fig:RD_main_both_MSSSIM}
\vspace{-3mm}
\end{figure}
\begin{figure}[t]
\centering
\subfigure[LPIPS]{
\centering
\includegraphics[width=0.475\linewidth]{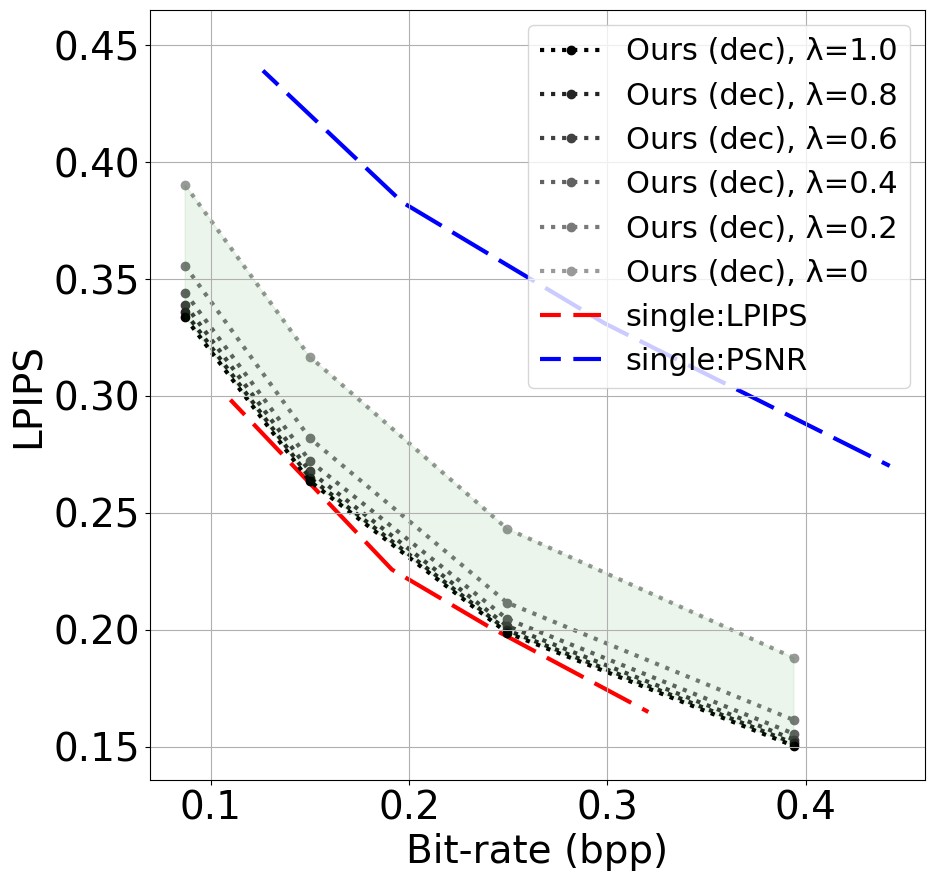}
}
\subfigure[PSNR]{
\centering
\includegraphics[width=0.455\linewidth]{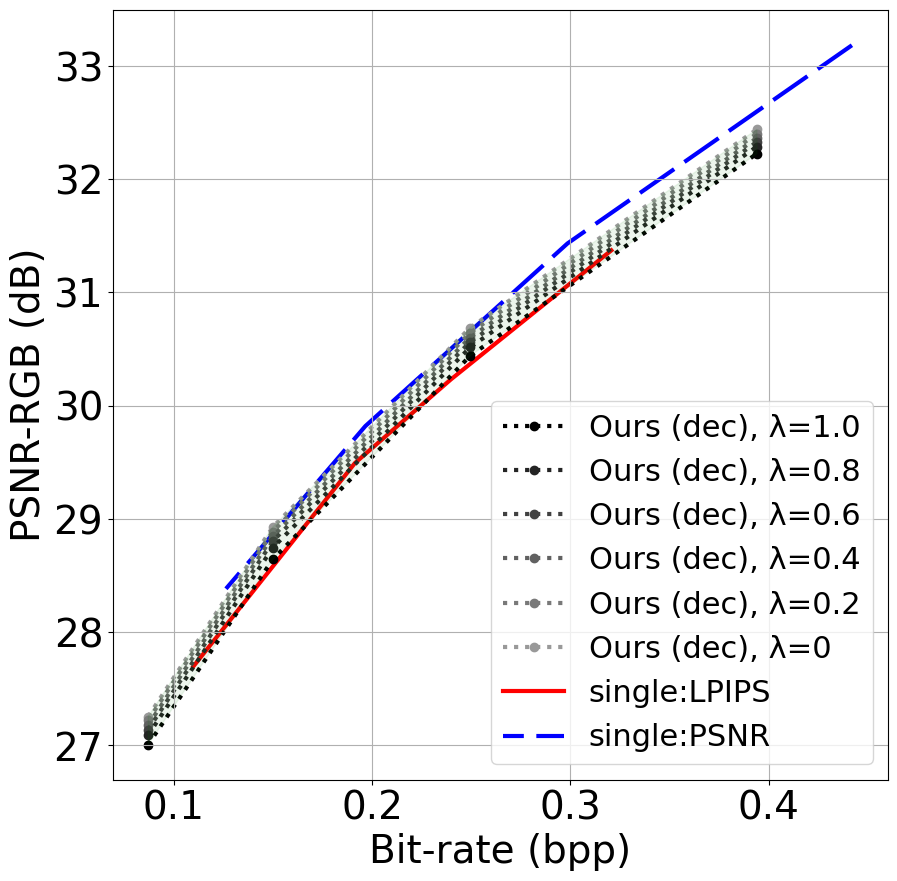}
}
\caption{Rate-distortion performance under (LPIPS, PSNR) pair with decoder-side only prompts. Green shaded area indicates the adaptive range of our method.}
\label{fig:RD_main_dec_LPIPS}
\end{figure}
\begin{figure}[t]
\centering
\subfigure[MS-SSIM]{
\centering
\includegraphics[width=0.46\linewidth]{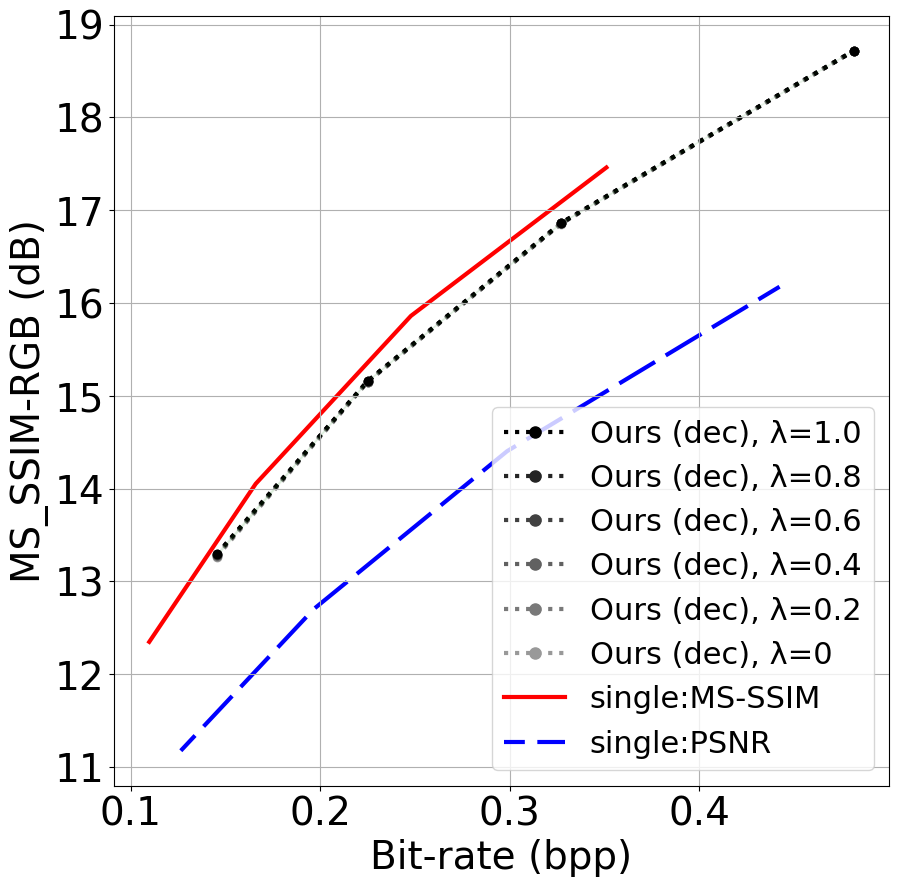}
}
\subfigure[PSNR]{
\centering
\includegraphics[width=0.46\linewidth]{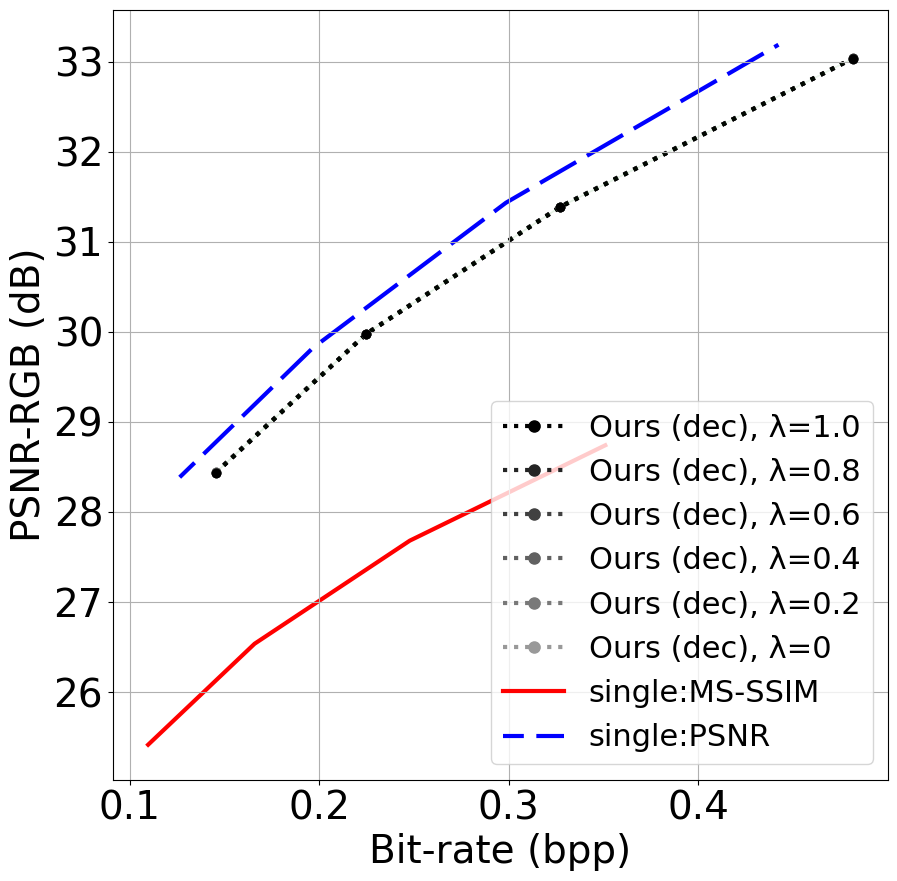}
}
\caption{Rate-distortion performance under (MS-SSIM, PSNR) pair with decoder-side only prompts. Green shaded area indicates the adaptive range of our method.}
\label{fig:RD_main_dec_MSSSIM}
\vspace{-3mm}
\end{figure}

As depicted in Fig.~\ref{fig:stb}, in W-MSA or SW-MSA, both $I_i$ and $P_i$ are arranged into windows of tokens, and the collocated windows are combined in the first step of the (S)W-MSA. Specifically, both the prompt and image tokens are involved in formulating the key and value matrices, whereas the query matrix is derived solely from the image tokens only. \textcolor{black}{The query matrix remains unchanged since only the output of image tokens is propagated and not that of prompt tokens.} In symbols, we have
\begin{equation}
\begin{aligned}    
    Q &= I\textbf{W}_Q\\
    K &= [P, I]\textbf{W}_K\\
    V &= [P, I]\textbf{W}_V,
\end{aligned}
\end{equation}
where $P, I$ collectively refer to the prompt and image tokens in the windows, and $[\cdot,\cdot]$ denotes concatenation along the token dimension. $\textbf{W}_Q, \textbf{W}_K, \textbf{W}_V\in\mathbb{R}^{d\times d}$ are learnable weights projecting the input into query $Q\in\mathbb{R}^{N\times d}$, key $K\in\mathbb{R}^{M\times d}$, and value matrices $V\in\mathbb{R}^{M\times d}$, where $d$ is the number of channels of each token, $N$ is the number of query tokens, and $M=N+\frac{N}{4}$ is the number of key (or value) tokens. Then we perform canonical self-attention as follows:
\begin{equation}
    \label{eq:msa}
    \begin{aligned}
        \text{Attention}(Q,K,V) &= \text{Softmax}(QK^\top/\sqrt{d}+B)V, \\
    \end{aligned}
\end{equation}
where $B$ is the learned relative positional bias matrix.

\begin{figure*}[t]
\centering
\subfigure[(LPIPS, PSNR), the text below each image indicates its BPP/PSNR/LPIPS]{
\centering
\includegraphics[width=0.9\linewidth]{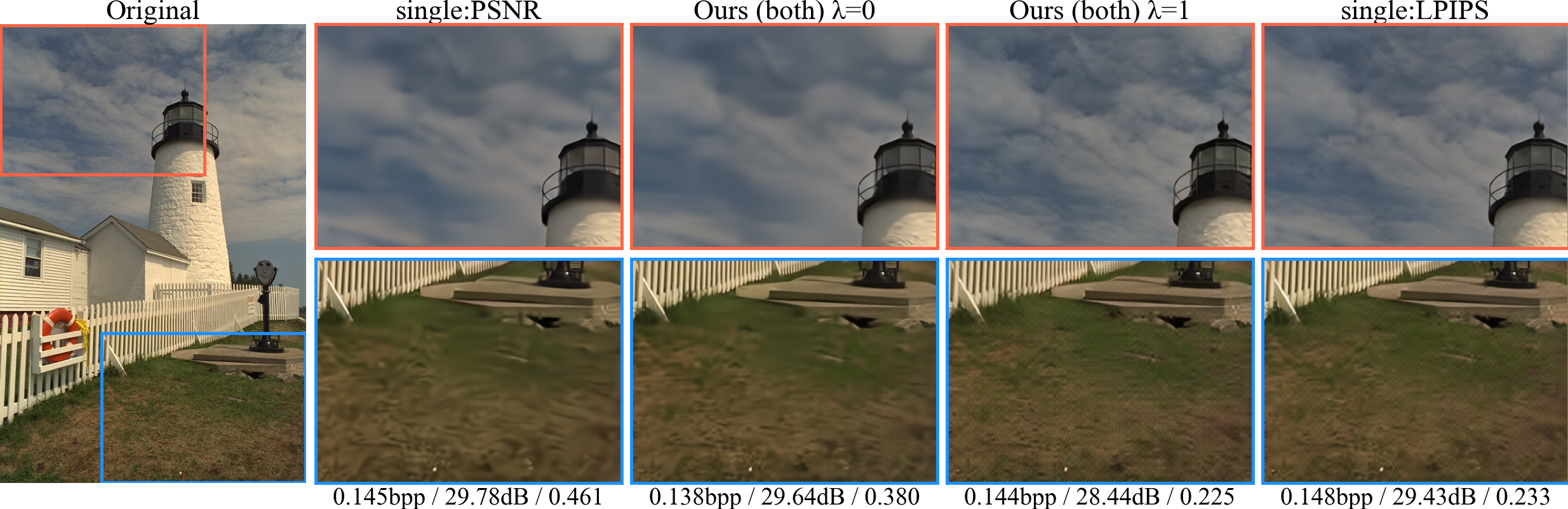}
\label{fig:Vis_both_lpips}
}
\subfigure[(MS-SSIM, PSNR), the text below each image indicates its BPP/PSNR/MS-SSIM]{
\centering
\includegraphics[width=0.93\linewidth]{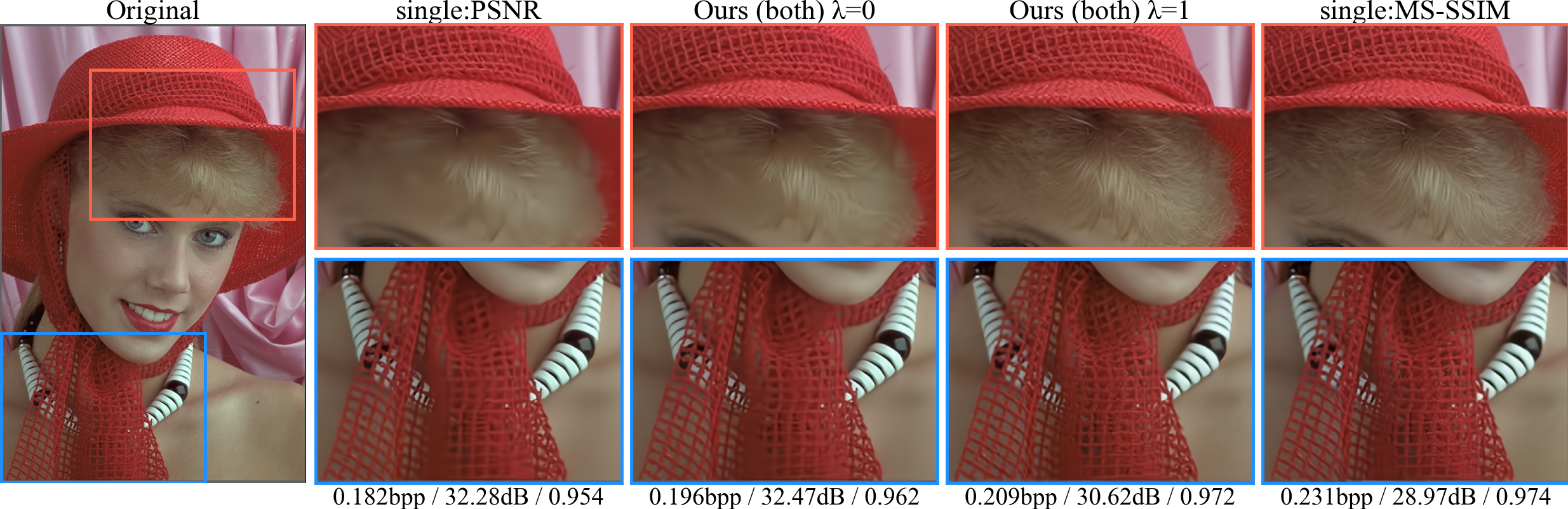}
\label{fig:Vis_both_msssim}
}
\caption{Subjective quality comparison of single-metric methods and our method with both-side prompts on two pairs of quality objectives. Zoom in for better visualization.}
\label{fig:Vis_both}
\vspace{-3mm}
\end{figure*}

\textcolor{black}{In this work, we explore two variants of the prompt-based conditioning methods for different application scenarios. The first one, both-side prompts, applies prompting to both the analysis and synthesis transforms by introducing $G_{enc}$ and $G_{dec}$. In this way, the latent $y$ is specialized for one specific trade-off $\lambda$. Although the $\lambda$ needs to be specified before encoding, this approach achieves better rate-distortion performance because of the fully-customized bitstream. The other variant conditions only the decoder, i.e. no $G_{enc}$, resulting in a shared latent $y$ that can be decoded with different $\lambda$s. However, it has certain limitations, including showing inferior rate-distortion performance or not being able to adapt on certain cases, which are discussed and compared with both-side conditioning in Section~\ref{sec:exp}.}

\subsection{Training Objective}
We follow the conventional training objective of learning-based image compression, which involves minimizing a rate-distortion cost, to train our model. In order to adapt between two selected quality objectives, we formulate the loss function $\mathcal{L}$ as
\begin{equation}
    \mathcal{L} = R_\lambda\underbrace{(-\log p(\hat{z}) -\log p(\hat{y}|\hat{z}))}_{Rate}+ \underbrace{\lambda d_A+(1-\lambda) d_B}_{Distortion},
\label{eqs:loss}
\end{equation}
where $d_A, d_B$ denote the distortion terms corresponding to two selected quality objectives, respectively, and $R_\lambda$ controls the trade-off between distortion and rate. When $\lambda$ is set to one, the distortion $d_A$ of the quality objective A dominates the distortion term, resulting in the optimization for the quality objective A only, and vice versa. During training, $\lambda$ is uniformly sampled so that the model is able to adapt the coding process to variable image quality objectives.

\section{Experiments}
\label{sec:exp}

\begin{figure*}[t]
\centering
\subfigure[(LPIPS, PSNR), the text below each image indicates its BPP/PSNR/LPIPS]{
\centering
\includegraphics[width=0.99\linewidth]{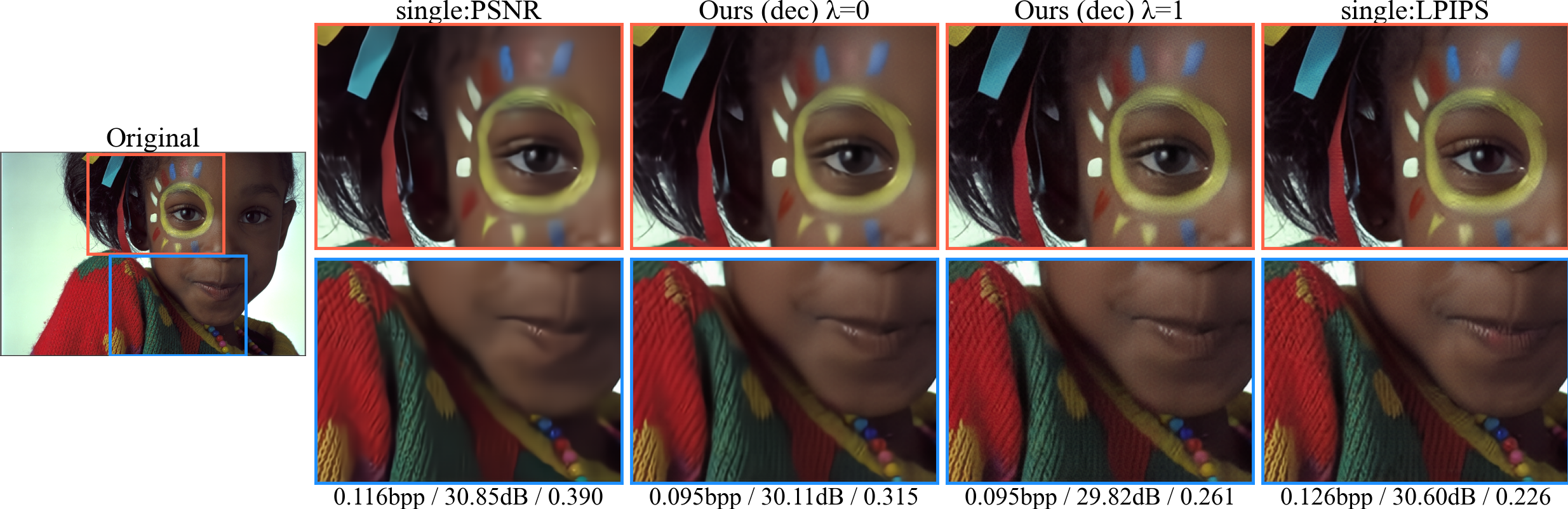}
\label{fig:Vis_dec_lpips}
}
\subfigure[(MS-SSIM, PSNR), the text below each image indicates its BPP/PSNR/MS-SSIM]{
\centering
\includegraphics[width=0.99\linewidth]{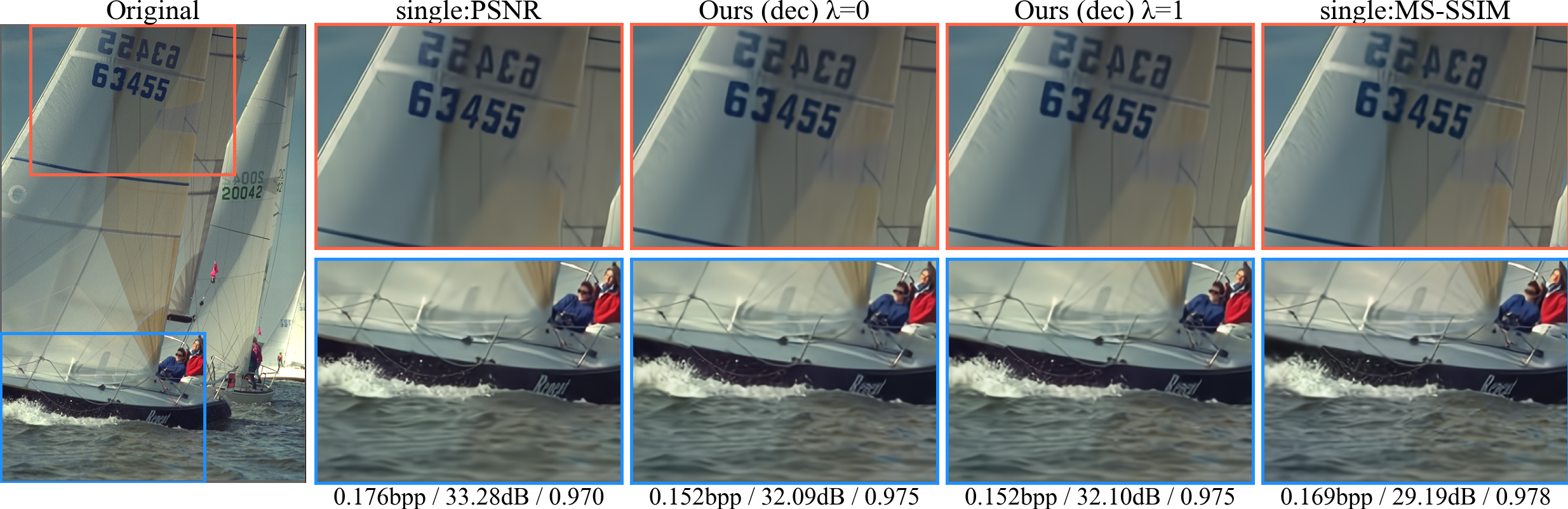}
\label{fig:Vis_dec_msssim}
}
\caption{Subjective quality comparison of single-metric methods and our method with decoder-side prompts on two pairs of quality objectives. Zoom in for better visualization.}
\label{fig:Vis_dec}
\vspace{-3mm}
\end{figure*}

\subsection{Settings}
\noindent\textbf{Quality Objectives.}
In this paper, the two quality objectives selected for adaptation are commonly-used image quality metrics. Specifically, we demonstrate the effectiveness and generalization of our proposed method on two selected quality objective pairs, (LPIPS, PSNR) and (MS-SSIM, PSNR).
Note that the two objectives in either pair exhibit distinct differences from each other. Optimizing for PSNR, which is simply minimizing the mean squared error between the reconstructed and original images, leads to blurry decoded images. On the other hand, MS-SSIM considers luminance, contrast, and structure distortions all at once, and LPIPS utilizes a pre-trained VGG network~\cite{simonyan2014very} to measure the image distortion in the feature domain. Both MS-SSIM and LPIPS tend to produce subjectively more pleasing results than PSNR. Each quality metric has its own distortion function $d$, as defined below:
\begin{equation}
\begin{aligned}
    d_{\text{PSNR}}(x,\hat{x}) &= 0.01\times\text{MSE}(x,\hat{x}),\\
    d_{\text{MS-SSIM}}(x,\hat{x}) &= 25\times(1-\text{MS-SSIM}(x,\hat{x})),\\
    d_{\text{LPIPS}}(x,\hat{x}) &= 2.5\times\text{LPIPS}(x,\hat{x}).\\
\end{aligned}
\label{eq:metric_distortion}
\end{equation}
By plugging these distortion functions into Eq.~\eqref{eqs:loss} as $d_A$ and $d_B$, we obtain the final training loss function for each objective pair. 

\noindent\textbf{Training.}
For both objective pairs, we first pre-train the base codec (i.e. $N_{enc}, N_{dec}, H_{enc}, H_{dec}$) by optimizing it only for PSNR following~\cite{lu2022transformer} on Flicker2W dataset~\cite{liu2020unified}. The resulting PSNR model is then combined with the prompt generation networks $G_{enc}$ and $G_{dec}$ for the next training stage, which jointly trains the entire model end-to-end using the loss function in Eq.~\eqref{eqs:loss} and Eq.~\eqref{eq:metric_distortion}. During training, we choose $R_\lambda=\{5.556, 2.857, 1.493, 0.769\}$, which is the reciprocal of the Lagrange multiplier used in~\cite{lu2022transformer} times 0.01. Note that $R_\lambda$ is associated with the rate term. The trade-off $\lambda$ is uniformly sampled between 0 and 1.

\noindent\textbf{Evaluation.}
We evaluate the performance of our proposed method on Kodak dataset~\cite{kodak}, which consists of 24 uncompressed images with resolution 512x768 or 768x512. The evaluation is done for both pairs of quality metrics (i.e. PSNR versus LPIPS, and PSNR versus MS-SSIM). When testing our proposed method, we choose $\lambda=\{0,0.2,0.4,0.6,0.8,1\}$ to evaluate the capability of the proposed method in fulfilling various quality objectives. 

\noindent\textbf{Baselines.}
We compare the performance of our proposed method with various baselines. Firstly, we train our base codec (i.e. $N_{enc}, N_{dec}, H_{enc}, H_{dec}$) by optimizing for a single quality metric, which is denoted as \textit{single:\{metric\}} (e.g. \textit{single:PSNR}). Note that for \textit{single:LPIPS}, we also include a MSE term during training to prevent instability when solely optimizing for LPIPS. In addition, we compare our prompt-based conditioning method with the existing methods: \textbf{SFT}~\cite{song2021variable} and \textbf{beta conditioning}~\cite{agustsson2022multi}. For a fair comparison, both conditioning methods are implemented on our Swin-Transformer base codec and trained following the same procedure as ours. In the case of SFT~\cite{song2021variable}, we adopt the conditioning networks from~\cite{song2021variable} and apply affine transformation to the output feature maps of every STB in both the encoder and decoder following \cite{song2021variable}. As for the beta conditioning~\cite{agustsson2022multi}, we employ the mapping function and projection to perform channel-wise shifting on the output feature maps of every Swin-Transformer layer in the decoder following~\cite{agustsson2022multi}. In order to compare with our both-side and decoder-side conditioning methods, we construct a variant of SFT with only the decoder-side condition. Likewise, we develop a variant of the beta conditioning method by applying it to both the encoder and decoder. 

\subsection{Performance Comparison for Both-side Prompts}
We first examine the compression performance of our proposed method with both-side prompts, where both $G_{enc}$ and $G_{dec}$ are utilized. For the (LPIPS, PSNR) objective pair, Fig.~\ref{fig:RD_main_both_LPIPS} compares the rate-distortion performance of our method with that of the single-metric methods. By adopting different $\lambda$ values, our method is able to adapt to a wide range of quality objectives (see the green shaded area). When $\lambda=0$, which translates to optimizing solely for PSNR, our method is able to reach the same performance as the single-metric method \textit{single:PSNR}. Notably, when $\lambda=1$, our method even outperforms marginally \textit{single:LPIPS}, which is due in part to the increased model capacity. These results suggest that our proposed method has the capability of fulfilling a wide range of image quality objectives with a single model, while performing comparably to or even better than the single-metric methods at the two extreme $\lambda$s. 


Similarly, Fig.~\ref{fig:RD_main_both_MSSSIM} shows that our method is capable of effectively trading off MS-SSIM for PSNR. Similar to (LPIPS, PSNR), our method performs comparably or superior to \textit{single:PSNR} on PSNR with $\lambda=0$ and \textit{single:MS-SSIM} on MS-SSIM  with $\lambda=1$. This demonstrates the versatility of our proposed method, which is able to handle various quality objective pairs with ease and achieve consistently good performance. 


Fig.~\ref{fig:Vis_both} visualizes the decoded images produced by our method and the single-metric methods for both pairs. The variable quality objectives are also reflected on the reconstructed images. 
For LPIPS and PSNR (Fig.~\ref{fig:Vis_both_lpips}), the two methods with higher PSNR (\textit{single:PSNR}, Ours $\lambda=0$) have smoother images with fewer details, while the other two with low LPIPS (\textit{single:LPIPS}, Ours $\lambda=1$) shows clearer edges and more refined structures. However, the latter two, when examined closely, have some artificial patterns. This may be attributed to the characteristics of the LPIPS metric. With our proposed method, the undesirable patterns can be reduced through adjusting $\lambda$ to strike a balance between the two quality objectives, while the single-metric methods have to re-train the entire codec for different objectives. With MS-SSIM and PSNR (Fig.~\ref{fig:Vis_both_msssim}), the results are similar. It is observed that ours with $\lambda=0$ produces blurry reconstructed images with less distinct edges and details. In contrast, ours with $\lambda=1$ and \textit{single:MS-SSIM} both exhibit sharper and more visually pleasing reconstructed images despite their lower PSNR. 


\begin{figure}[t]
\centering
\subfigure[MS-SSIM]{
\centering
\includegraphics[width=0.46\linewidth]{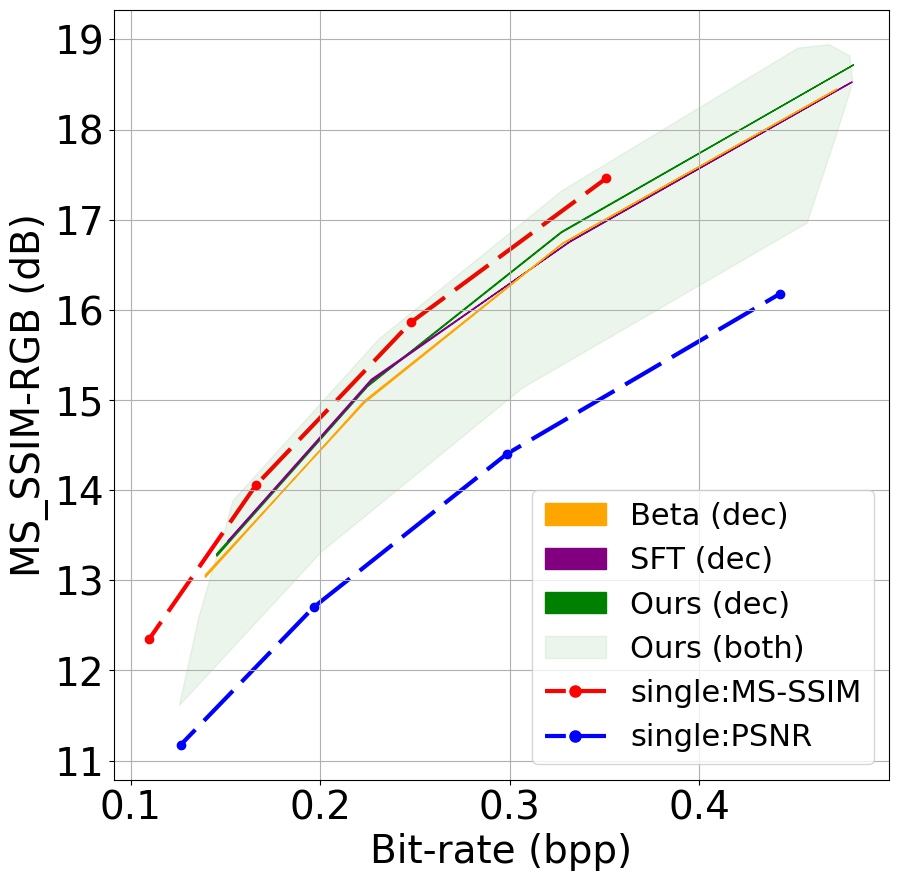}
}
\subfigure[PSNR]{
\centering
\includegraphics[width=0.46\linewidth]{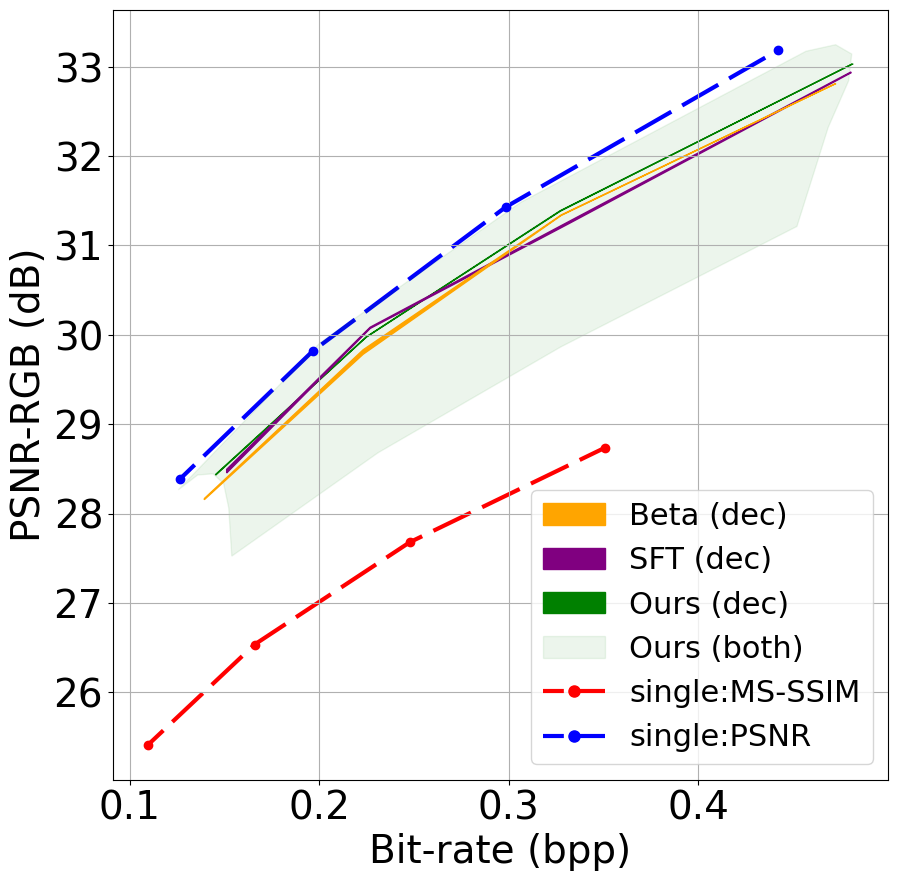}
}
\caption{Adaptive range comparison under (MS-SSIM, PSNR) pair with various conditioning methods.}
\label{fig:RD_comp_shade_MSSSIM}
\vspace{-3mm}
\end{figure}
\begin{figure}[t]
\centering
\subfigure[MS-SSIM]{
\centering
\includegraphics[width=0.46\linewidth]{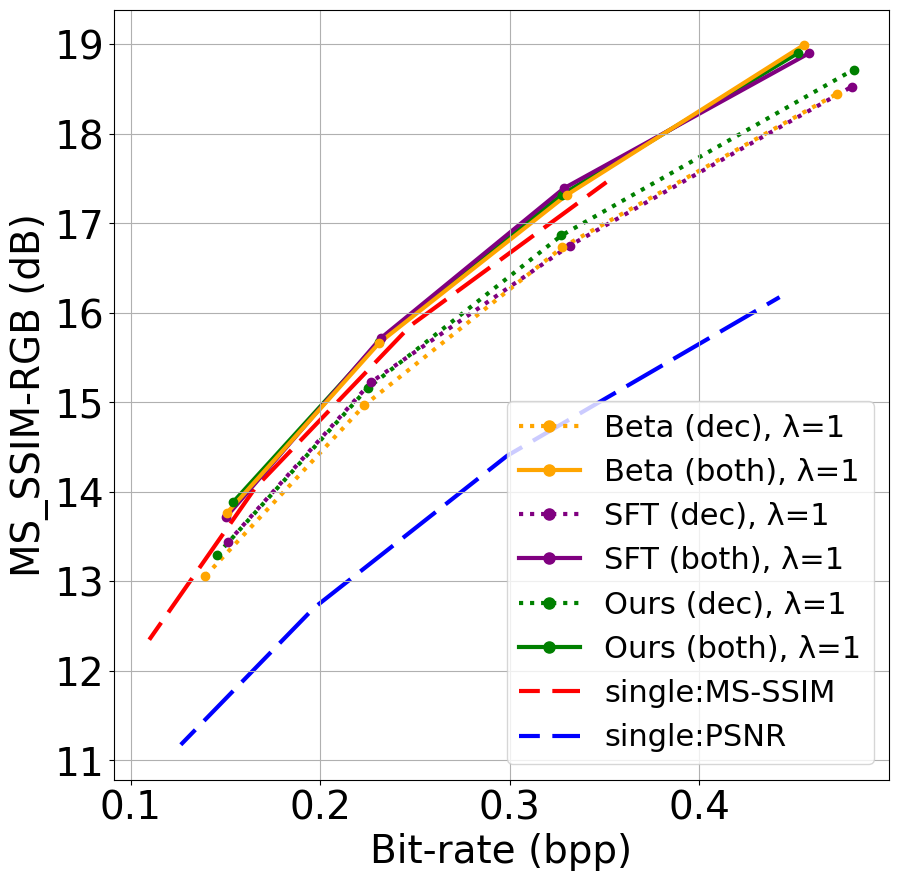}
}
\subfigure[PSNR]{
\centering
\includegraphics[width=0.46\linewidth]{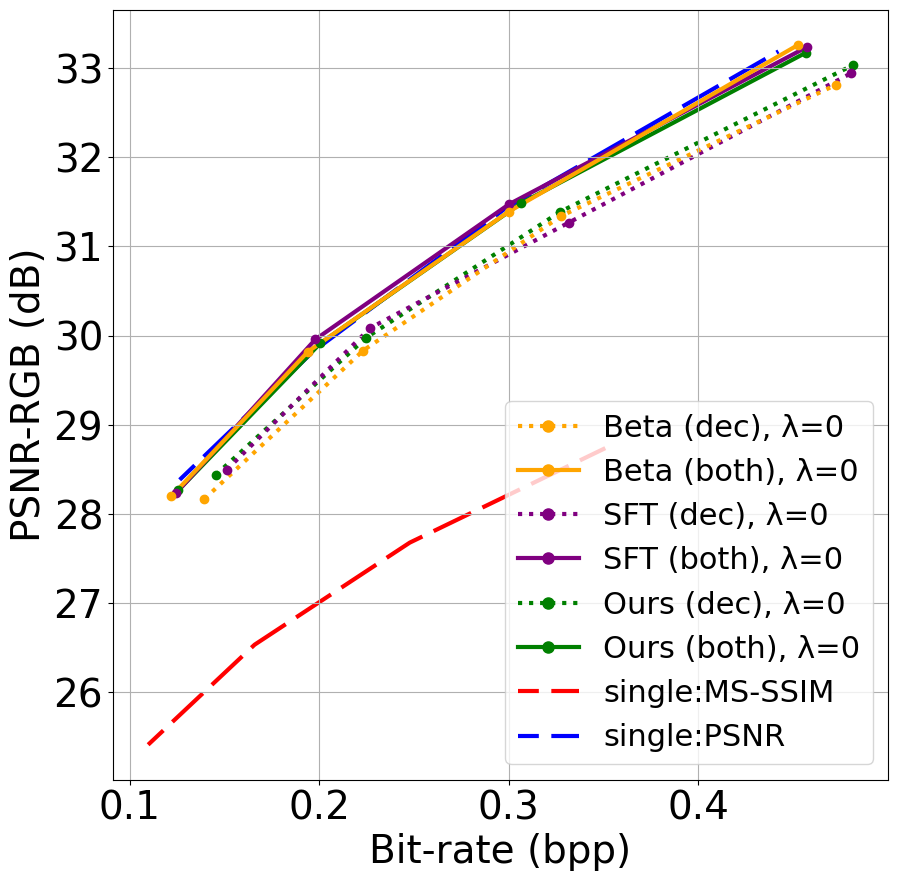}
}
\caption{Rate-distortion performance comparison under (MS-SSIM, PSNR) pair with various conditioning methods at $\lambda=1$ or $\lambda=0$.}
\label{fig:RD_comp_MSSSIM}
\end{figure}
\begin{figure}[t]
\centering
\subfigure[LPIPS]{
\centering
\includegraphics[width=0.475\linewidth]{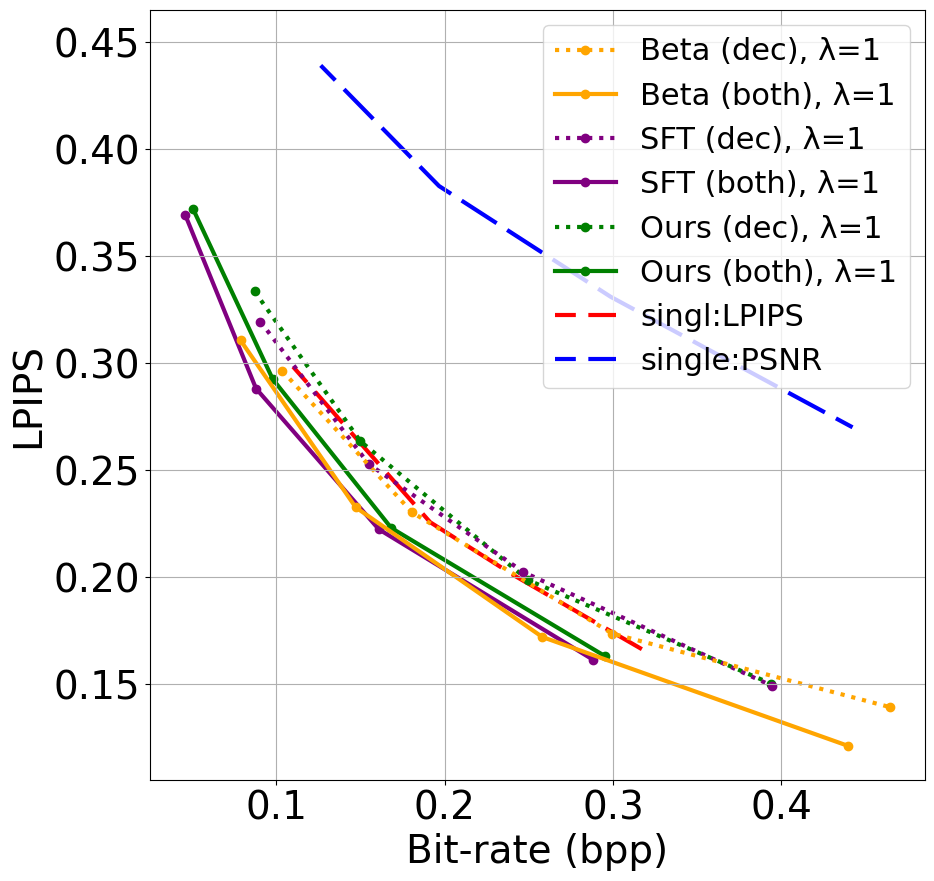}
}
\subfigure[PSNR]{
\centering
\includegraphics[width=0.455\linewidth]{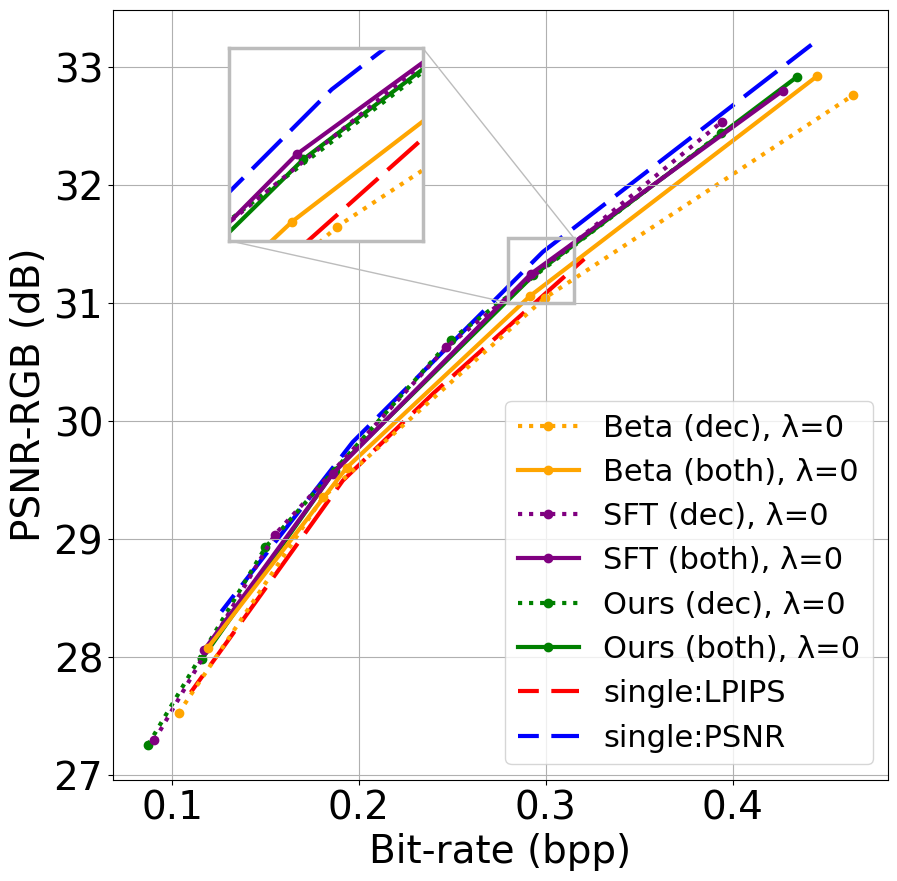}
}
\caption{Rate-distortion performance comparison under (LPIPS, PSNR) pair with various conditioning methods at $\lambda=1$ or $\lambda=0$.}
\label{fig:RD_comp_LPIPS}
\vspace{-4mm}
\end{figure}

\subsection{Performance Comparison for Decoder-side Prompts}
Next, we analyze the compression performance of our method with decoder-side prompts alone, where only $G_{dec}$ is utilized. Fig.~\ref{fig:RD_main_dec_LPIPS} presents the rate-distortion results of the (LPIPS, PSNR) objective pair. While our method is still able to adjust the reconstructed quality with different $\lambda$ values, the range of adaptation is relatively smaller in comparison with the both-side results shown in Fig.~\ref{fig:RD_main_both_LPIPS}. At the two extreme $\lambda$ values, the adapted results barely match the rate-distortion performance of the single-metric methods. This observation may be attributed to the removal of $G_{enc}$, thereby producing a single bitstream that needs to be decoded at different $\lambda$ values.


Fig.~\ref{fig:RD_main_dec_MSSSIM} presents the rate-distortion results on the (MS-SSIM, PSNR) objective pair. Contrary to the previous results with both-side prompts, the rate-distortion performance remains somewhat fixed regardless of the given $\lambda$, with all six curves coinciding with each other. This indicates that when relying only on the decoder for adaptation, our method is unable to trade off PSNR for MS-SSIM. This result suggests that there exist some metric pairs where the metrics involved are incompatible with each other in the sense that when a single bitstream is used, the decoder-side only adaptation is not able to fulfill variable quality objectives. We further verify this through experiments with the other conditioning methods in the next section. 

Fig.~\ref{fig:Vis_dec} presents the subjective quality comparison of decoded images obtained through the single-metric methods and our method on both pairs of metrics. For LPIPS and PSNR, our method with $\lambda=1$\textcolor{black}{, which corresponds to optimizing for LPIPS,} displays richer details \textcolor{black}{but also exhibits more prominent artificial patterns}, the result of which is similar to \textit{single:LPIPS} (Fig.~\ref{fig:Vis_dec_lpips}). In contrast, $\lambda=0$ and $\lambda=1$ produce nearly identical images on MS-SSIM and PSNR (Fig.~\ref{fig:Vis_dec_msssim}). 

\subsection{Comparison with Competing Methods}
In this section, we compare our proposed method with the other conditioning methods, namely SFT~\cite{song2021variable} and beta conditioning~\cite{agustsson2022multi}. First, we focus on the failure case with decoder-side prompts on MS-SSIM and PSNR, and test both baselines under the same setting of decoder-side only conditioning. Fig.~\ref{fig:RD_comp_shade_MSSSIM} illustrates the effective range of adaptation. It is apparent that all methods using decoder-side only conditioning are unable to display any substantial adaptation range, which is a direct contrast to our variant with both-side prompts. This further indicates that the failure to adapt between MS-SSIM and PSNR lies in the combination of these objectives rather than the proposed conditioning method. 

Fig.~\ref{fig:RD_comp_MSSSIM} and Fig.~\ref{fig:RD_comp_LPIPS} further compare the rate-distortion performance of all the methods at $\lambda=1$ and $\lambda=0$ under the both-side and decoder-side settings. For the (MS-SSIM, PSNR) objective pair, Fig.~\ref{fig:RD_comp_MSSSIM} shows that while all three methods with the both-side conditioning show similar performance on both metrics, our proposed prompt-based method outperforms SFT~\cite{song2021variable} and the beta conditioning~\cite{agustsson2022multi} when using the decoder-side only conditioning. This suggests that prompting offers a better means to condition Swin-Transformer-based models. As for the (LPIPS, PSNR) metric pair, Fig.~\ref{fig:RD_comp_LPIPS} shows that the beta conditioning performs the best in terms of LPIPS; however, it exhibits a significant drop in PSNR under the both-side and decoder-side settings. On the other hand, although SFT achieves a slight gain over our method in terms of LPIPS with the both-side prompts, we note that SFT introduces considerably higher complexity than our method, adding nearly 15 million model parameters as opposed to 6 million with our scheme. Overall, our proposed method strikes a better balance among the adaptation capability, rate-distortion performance, and complexity.

\section{Conclusions}
In this paper, we propose a practical yet efficient solution for adapting to variable quality objectives in image compression, which is achieved through inserting model-generated prompts into the Swin-Transformer based codec. It offers superior performance compared to existing methods and demonstrates consistent results across different quality objective pairs. We additionally explore two variants of our proposed method and compare them in terms of performance and applications. The ability to trade off one quality objective for another using a single model, without the need for retraining, makes our method highly versatile and applicable to a wide range of scenarios.

\section*{Acknowledgement}
{\textcolor{black}{This work is supported by National Science and Technology Council, Taiwan under Grants NSTC 111-2634-F-A49-010- and MOST 110-2221-E-A49-065-MY3, MediaTek, and National Center for High-performance Computing.}}

\bibliographystyle{ieeetr}
\bibliography{refs}

\end{document}